\documentclass[10pt]{article}

\usepackage{amsmath}
\usepackage{amssymb}

\usepackage{graphicx}

\usepackage{cite}

\usepackage{color} 
\usepackage{url}
\usepackage{subfigure}
\usepackage{ulem}

\usepackage{setspace} 
\usepackage[labelfont=bf,labelsep=period,justification=raggedright]{caption}

\makeatletter
\renewcommand{\@biblabel}[1]{\quad#1.}
\makeatother

\begin{document}

\begin{flushleft}
{\Large
\textbf{Population Structure Analysis of an Inbred Population using Quantitative Shape Phenotyping from Stereo Retinal Photographs}
}
\newline
\\
Li~Tang$^1$, 
Michael~D.~Abr\`{a}moff$^{2,\ast}$
\newline
\\
\bf{1} Department of Ophthalmology and Visual Sciences, University of Iowa, Iowa City, IA, USA
\\
\bf{2} Department of Ophthalmology and Visual Sciences, Stephen A Wynn Institute for Vision Research, Department of Biomedical Engineering, and Department of Electrical and Computer Engineering, University of Iowa, Iowa City, IA, USA; Iowa City Veterans Administration Medical Center, Iowa City, IA, USA
\newline
\\
$\ast$ Correspondence: michael-abramoff@uiowa.edu
\end{flushleft}

\newpage

\section*{Abstract}

The population structure of an inbred population of 781 people on Norfolk Island in the Pacific, 318 of which are descendants of the original Mutineers of the Bounty, is analyzed phenotypically using shape from stereo retinal fundus photographs. Three-dimensional optic nerve head (ONH) shape is reconstructed from stereo pairs by a multi-scale stereo matching algorithm. Using deep neural network, the shape of ONH, which is under genetic control, is decomposed into a set of hierarchical features through self-taught learning. Features captured at different levels are selected according to their discriminant power in identifying the two populations. The prediction accuracy is evaluated with stratified cross validation. Given the selected feature set, individuals are grouped into k hierarchical clusters and cluster membership fractions are determined for k=2,3,4,5,6,7. Population structure analysis on the basis of phenotypes through image analysis allows heritability and linkage analysis, including founder effects from English and Polynesian ancestors, potentially leading to new genetic risk factors for glaucoma and other ONH-related eye diseases.

\section{Introduction}

Population structure analysis is a method to quantify similarities and dissimilarities between individuals in a population, and is essential for the discovery of the role of genetic risk factors in disease, as the distribution of genes in a population is highly associated with population structure. Usually, population structure analysis uses genotypes to group individuals into clusters that are genetically closer \cite{Science}, and these have been shown to be highly associated with self-reported ancestry, native language, geographical location of residency, genealogy, and other factors of gene expression.

Because gene expression is modulated by the environment, and because the past environment is usually unknown, causing more noisy data, population structure analysis on the basis of phenotype (expression of the genetic background of the individual) is more variable. However, because the phenotypic variation can be associated with putative genotypes, it is highly useful to discover new genes. Especially, endophenotypes (varying phenotypes within individuals with the same disease) can help discover specific genes in multi-genetic diseases.

However, phenotypic population structure analysis is either based on the measurements of single or low dimensional gene expression such as height, weight, blood pressure etc., or on very laborious manual, multidimensional measurements, including anthropometrics such as skull phenotypes, demratoglyphs or dental traits.  \cite{admixture}. Complex structures, such as human skull observed as craniofacial morphology, have been studied in population after high levels of admixture \cite{admixture}. Results indicate that it fits the theoretical expectations of quantitative genetics and is predictive of population structure and history. Phenotypic population structure analysis based on more complex, high-dimensional phenotypes obtained from images, such as facial structure or retinal morphology from retinal images, has not been attempted.

The morphology of the optic nerve head (ONH) plays a critical role in diagnosis and management of many optic neuropathies \cite{MDA}. Recent evidence has suggested that its morphology, especially its diameter (disc size), as well as its general shape forms phenotypes that are heritable \cite{Hewitt:IOVS}. However variations in ONH morphology, such as neuroretinal rim slope and curvature of the retinal nerve fiber layer (RNFL), may contain complex phenotypes that are difficult to access and quantify for human experts. Description and quantification of the three dimensional shape of the ONH and statistically correlate the findings with genes have been a challenging topic.

Different imaging techniques have been developed to monitor the ONH shape, such as stereo retinal photography and optical coherence tomography (OCT) \cite{Huang:OCT, Schuman:OCT}. Stereo retinal photography, as a traditional noninvasive imaging technique, has the advantages of being low cost and easily accessible. However its interpretation has primarily been subjective, qualitative and imprecise. OCT has recently become the method of choice to quantify the topography of the ONH. However it is not available in most clinics, and more importantly, for years and even decades may not yet offer longitudinal dataset easily collected from large populations.

Shape-from-stereo (SFS) made it possible to derive three-dimensional shape from stereo images, which has been focus of attention in computer vision community for decades \cite{Marr:Science,Szeliski:taxonomy,Brown:Advances}. It involves estimation of three-dimensional shape or disparity of correspondences using two images of the same scene from slightly different view angles. Specifically, a robust measurement, represented as disparity or shape map of ONH, can be derived from stereo retinal photographs in the presence of commonly observed spatially-varying reflectance, blur, noise, and low contrast or feature density, because of limitations on the amount of illumination for patient safety and imperfections of the optics of the eye.

Complex phenotypes encoded in the shape map of ONH are quantified through a deep neural network, as it represents increasingly complex features in a hierarchical architecture by integrating a successive stages of processing, which mimic the function of the visual system and the abstraction of human perception. Deep learning of complex features from large datasets and its performance analysis attract more and more attentions for its potentials of improving our understanding of the visual system itself, and eventually, the mechanics of generating representation of increasing levels of abstraction by the brain \cite{KoHo:Science}.

In this study, we determined population structure of a representative sample of an inbred population of Norfolk Island in the Pacific, of which half the population can trace their ancestry back to the Mutiny of the Bounty mutineers and the Tahitian wives they took with them, originally to Pitcairn island, automatically form morphologic phenotypes obtained from retinal stereo fundus images. This highly inbred population is ideal to test the expectations of quantitative genetics for admixture events and gene flow. We use a robust multi-scale stereo correspondence approach to estimate the shape of ONH from optic disc photographs. The outcome disparity maps, or topographic maps of ONH are further condensed by training a deep neural network to quantify this three-dimensional structure within a hierarchical feature space. We performed morphological quantification of the ONH from stereo photographs, determined phenotypes expressed in the population, measured the structure of the population, and validated our approach with a single truth - whether the individual could trace ancestry back to the Mutineers or not.

\section{Subjects and Methods}

\subsection{Related Work}

Suppose a pair of photographs is captured simultaneous using a fixed-stereo base fundus camera, there are disparities along horizontal coordinates between pairs of matched points $(x,y)\in{I}_1$ and $(x+d(x,y),y)\in{I}_2$. Magnitudes of horizontal disparities of all densely matched correspondences $d(x,y)$ form a disparity map $D$ reflecting the shape of imaged objects, in our case ONH. We previously designed a robust stereo matching algorithm for fundus photographs, and validated the methodology quantitatively by comparing with the same structure obtained from spectral domain optical coherence tomography (SD-OCT) scans \cite{Tang:PAMI}. The root mean squared (RMS) difference between normalized structures was $15.9\pm{8.8}\%$ of the cup depth \cite{Li:SPIE,Tang:PAMI}, demonstrating the shape estimate from stereo retinal images faithfully reproduced the topography of the ONH as confirmed by SD-OCT volumes.


We observed that quantitative phenotyping of the ONH shape from stereo retinal photographs leads to phenotypes that can be measured and are largely under genetic control \cite{Li:IOVS}. The validation was performed using a random sample of 172 subjects (344 eyes, 45 pairs of monozygotic [MZ] and 41 dizygotic [DZ] twins) from the Twins Eye Study in Tasmania. Results show that inherited three-dimensional ONH shape parameters of twins can be quantified successfully and approximately 80\% of the variability in ONH shape parameters is determined genetically.

\subsection{Learning ONH Shape in Hierarchical Feature Space}

Given two topographic maps of ONH, a comparison between underlying shapes or patterns with respect to a distance metric is only meaningful after they have been transformed into a feature space that captures discriminative features of the anatomical structure itself and stay invariant to other irrelevant factors, such as translation, tile of the retinal surface or other small depth perturbations due to noise introduced in computation. In order to perform such statistical shape analysis, disparity maps are decomposed into a hierarchical feature space by training a deep neural network with a symmetric structure, consisting both an encode and a decoder \cite{Hinton:Science}. The features captured at sequential layers of such a stacked autoencoder (SAE) form the desired vector characterizing distinct ONH shape and the overall variability. By representing disparity map in such a transformed feature space, the similarity between two shapes can be measured quantitatively in terms of distance metrics of the associated feature vectors.

A set of three dimensional shapes of ONH, represented by $N$ disparity maps $\{\textbf{D}_1,\textbf{D}_2,\ldots\textbf{D}_N\}$ of $h\times{w}$ pixels, are re-arranged as $N$ long vector $\textbf{x}_k$ of length $h\times{w}$, where $h$ and $w$ represent the height and the width of the pre-processed disparity map. Each vector $\textbf{x}_k$ in the set is rescaled to a range of $[0,1]$. Maps from a left eye are flipped to match the ONH shape of a right eye. Stacked autoencoder \cite{Bengio:NIPS,Bengio:Review} is trained to project this high dimensional shape space to layers of hierarchical features by a set of parameters, i.e. weights $\textbf{w}_{ij}$ and bias $\textbf{b}_i$ of the deep neural network.

In the multi-layer network, output vector $\textbf{a}_j$ of layer $j$ is obtained by a linear combination of output vector $\textbf{a}_i$ of the previous layer $i$, following a non-linear transformation:
\begin{eqnarray}
\label{Eq:activation}
\textbf{a}_j=sigm(\textbf{w}_{ij}\textbf{a}_i+\textbf{b}_i)
\end{eqnarray}

\noindent where non-linear activation $sigm(\cdot)$ represents the sigmoid function $sigm(\textbf{x})=1/(1+e^{-\textbf{x}})$. Given this network structure, one unit at layer $j$ is determined by multiple units at layer $i$ and the strength of weights connecting them, which means the learning is essentially a process to adapt parameters to respond to coactivation of certain patterns and be selective to structures arose progressively at different spatial scales \cite{Seung:Nature}.

\begin{figure}[!ht] 
\begin{center}
\subfigure[]{\includegraphics[height=1.5in]{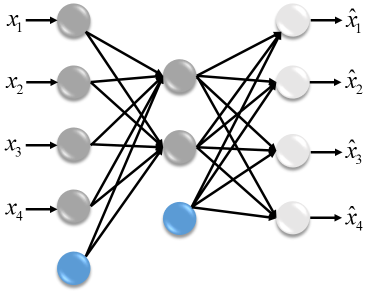}
\label{fig:network}}\hspace{8mm}
\subfigure[]{\raisebox{5mm}{\includegraphics[height=1.2in]{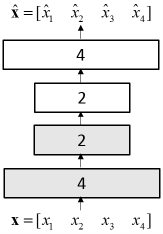}
\label{fig:block}}}
\end{center}
\caption{Network structure of a sparse autoencoder and its simplified representation.}
\label{fig:netexample}
\end{figure}

Fig.~\ref{fig:netexample} illustrates the network structure of a simple sparse autoencoder with 4 input units and 2 hidden units, where the gray units represent encoder layer, the white units represent decoder layer, and the blue units represent the bias terms. The weights are denoted as the connections between the input units and the hidden units. The same network structure can be represented in a simplified form as the one shown in the right panel. By minimizing the differences between input vector $\textbf{x}$ and output vector $\hat{\textbf{x}}$ of a set of training samples, the network is able to encode the information contained in a 4-dimensional vector into a 2-dimensional vector with a set of learned parameters (i.e. weight and bias). Features derived from such an autoencoder through self-taught learning turn out to improve classification performance for various pattern recognition tasks \cite{AndrewNg:Selftaught}. Another main advantage of this approach is that it requires only unlabeled data, which is more easily available for most of real applications.

\begin{figure}[!ht] 
\begin{center}
\subfigure[]{\includegraphics[height=0.8in]{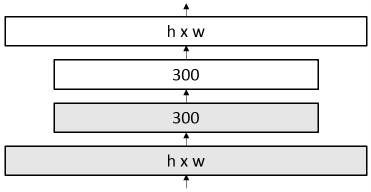}
\label{fig:stage1}}
\subfigure[]{\includegraphics[height=0.8in]{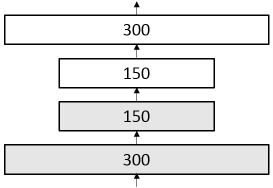}
\label{fig:stage2}}
\subfigure[]{\includegraphics[height=0.81in]{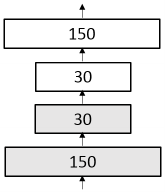}
\label{fig:stage3}}
\end{center}
\caption{Pretraining is performed by training each of the sparse autoencoder sequentially in the stack.}
\label{fig:pretraining}
\end{figure}

The training of stacked autoencoder involves pretraining each of the cascaded sparse autoencoder, unrolling them into a stacked encoder and a stacked decoder, and fine tuning parameters associated with the re-organized network structure \cite{Hinton:Science}. The process of pretraining each of the stacked autoencoder separately provides reasonable initialization of parameters for the final fine tuning step. The three stages are illustrated with an example network setting in Fig.~\ref{fig:pretraining}, where gray boxes indicate encoder layers and white boxes decoder layers.

In this example, three sparse autoencoders are pretrained using batch gradient descent by minimizing cost function $J(\textbf{w}, \textbf{b})$ with respect to parameter $\textbf{w}_{ij}$ and $\textbf{b}_i$ \cite{AndrewNg:ICML,AndrewNg:NIPS}:
\begin{eqnarray}
\label{Eq:cost}
J(\textbf{w}, \textbf{b})=\frac{1}{m}\sum_{i=1}^{m}\big(\frac{1}{2}||\textbf{a}^{(i)}-\hat{\textbf{a}}^{(i)}||^2\big)+\frac{\lambda}{2}\sum_l\sum_i\sum_j(\textbf{w}_{ij}^{(l)})^2 \nonumber\\
+\beta\sum_j\big[\rho{log}\frac{\rho}{\hat{\rho}_j}+(1-\rho)log\frac{1-\rho}{1-\hat{\rho}_j}\big]
\end{eqnarray}

\noindent where $\lambda$, $\beta$ and $\rho$ are constants involved in the weight decay term and the sparsity penalty term. $\hat{\rho}_j$ denotes the average activation of hidden unit $j$ of all $m$ training samples. The cost $J(\textbf{w}, \textbf{b})$ consists of three terms. The first term is the mean squared error between input data $\textbf{a}$ and recovered data $\hat{\textbf{a}}$. The second term is a regularization term (i.e. weight decay term) to prevent overfitting. The third term is a sparsity penalty term to restrict the average activation of each hidden neuron, which is crucial for a parts-based representation for perception \cite{Seung:Nature}. Cost function $J(\textbf{w}, \textbf{b})$ is minimized by computing its partial derivatives with respect to $\textbf{w}$ and $\textbf{b}$ using back propagation algorithm \cite{Duda:Book}.

\begin{figure}
\centering
		\includegraphics[height=4.5cm]{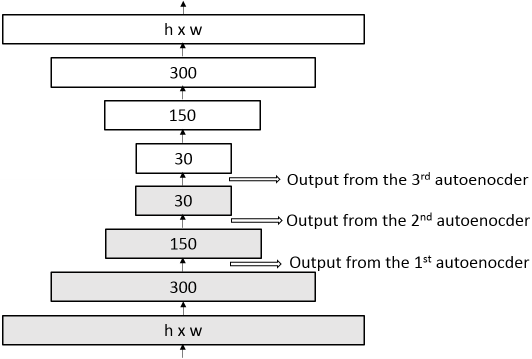}
\caption{Network structure to fine tune the stacked autoencoder.}
\label{fig:finetune}
\end{figure}

Given the parameters learned at the pretraining step, the stacked autoencoder is initialized accordingly in the \emph{unrolled} network depicted in Fig.~\ref{fig:finetune}. Each of the pretrained encoder is cascaded in sequence following their counterpart cascaded decoders in a symmetric fashion.

Fine tuning is performed by minimizing the error between the original disparity maps and the recovered disparity maps, which is achieved by forward propagation of the original maps through the cascaded layers of encoder and then the decoder. The derivatives for gradient descent is obtained by back propagation of the error derivatives through the cascaded layers of decoder and then the encoder. The resulting network thus learns the variability among the training samples and decomposes them into a set of basis components, which is encoded in the parameter space of $\textbf{w}$ and $\textbf{b}$. The outputs from each of the stacked autoencoder (Fig.~\ref{fig:finetune}) express the magnitude of each of those basis components that compose of the current input pattern. That means they form a set of feature vector characterizing distinct ONH shape, i.e. the network is trained to learn a better representation of ONH shape with a set of hierarchical features. The extracted features made it possible to perform quantitative analysis of topography of shape with a minimum loss of information present in the original structure.

Principal component analysis (PCA) \cite{Eigenfaces} is another widely used alternative to represent the variability of data, where each disparity map can be represented as a linear combination of a set of eigen disparity maps. Compared with PCA, SAE provides a non-linear representation of data and has the advantage of learning \emph{spatially localized} features without the constraint that the basis components, i.e. the eigen disparity maps have to be mutually orthogonal \cite{Olshausen:Nature}. While PCA is ideal to represent and compress data with a general statistical distribution, such as a Gaussian cloud, SAE is more appropriate to capture spatially localized features or structures, such as those described by natural images or structures. A performance comparison between SAE and PCA is present in Section~\ref{sec-result}.

\subsection{Hierarchical Clustering of the Phenotypic ONH Shape}

In this study, we investigated complex ONH shape phenotypes using a unique inbred population. Its origins can be traced back to 1789.

\paragraph{Historical and demographical background.}

The Mutiny of the Bounty occurred in April 1789, and the original mutineers (Fletcher Christian, the Bounty's acting Lieutenant and 7 other English males) and 12 females from Tahiti (Polynesian ancestry) fled to Pitcairn Island in the Pacific Ocean \cite{Legacy}. In 1856, 194 inhabitants were relocated to Norfolk Island, and today almost 5700 individuals on Norfolk Island claim mutineer ancestry.

In 2008, 781 subjects, among which 343 are direct descendants of the mutineers, underwent retinal imaging, DNA genotyping and other biometrics \cite{Mackey:NIES}. The combination of a small number of original founders with diverse ancestries can be considered a natural experiment for admixture studies and present a unique population for the investigation of complex phenotypes. We test the hypothesis that the phenotypic ONH shape reflects the admixture process among this population. The contributions of the Pitcairn and the Norfolk population to their gene pool are expected to be different because of biological and historical factors.

Given the feature space derived from SAE, variations of ONH shape are grouped to different subsets by k-means clustering, which partition the shape set into $k$ clusters and each shape belongs to the cluster with the nearest mean. The number of the cluster $k$ varies continuously from 2 to 6. The distance metric between two feature vectors $\textbf{x}_r$ and $\textbf{x}_s$ is measured by their cosine distance:
\begin{eqnarray}
\label{Eq:cosDistance}
d(\textbf{x}_r,\textbf{x}_s)=1-\frac{\textbf{x}_r{}\textbf{x}_s'}{\sqrt{\textbf{x}_r'{}\textbf{x}_r}\sqrt{\textbf{x}_s'{}\textbf{x}_s}}
\end{eqnarray}

\noindent The within-cluster sum of squared distance is minimized for a desired partition:
\begin{eqnarray}
\label{Eq:kmeans}
min\sum^{k}_{i=1}\sum_{x_j\in{i}}d^2(\textbf{x}_j,\mu_i)
\end{eqnarray}

\noindent With the cluster number $k$ varying continuously, hierarchical clustering is achieved by grouping data over a variety of levels and creating a cluster tree or dendrogram, which provides observation of the multilevel hierarchy and potential structures present in the dataset, as clusters at one level are joined as clusters at the next level. It also allows determination of the level of clustering that is most appropriate to distinguish certain structures.

\section{Results}
\label{sec-result}

\subsection{Data Collection}

722 of the 781 subjects, including 1444 pairs of stereo retinal photographs of both eyes are analyzed with given labels indicating either from the Pitcairn Island or from the Norfolk Island. Among them 318 subjects (44\%) have mutineer descent and the rest (404 subjects) do not. As the number of this labeled dataset is fairly limited, another unlabeled dataset - the Australian Twin dataset is included for self-taught learning of general variability of ONH shape, which contains 4387 pairs of stereo fundus photographs. Note that this dataset is included only for the purpose of training SAE and identifying hierarchical features extracted from disparity map in general.

Stereo pairs were excluded (781-722) if their labels were unavailable or there existed clear error in three-dimensional ONH shape estimate by visual inspection. This may happen when two images of the stereo pair have obviously illumination artifacts, noise, or noticeably out of focus blur.

\subsection{Image Preprocessing}

\begin{figure}[!ht] 
\begin{center}
\subfigure[]{\includegraphics[height=1.2in]{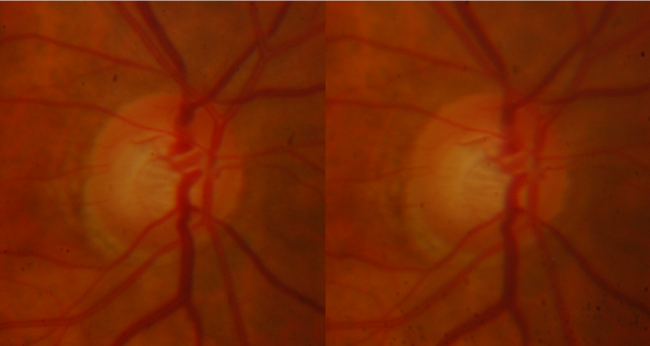}
\label{fig:origFundus}}
\subfigure[]{\includegraphics[height=1.2in]{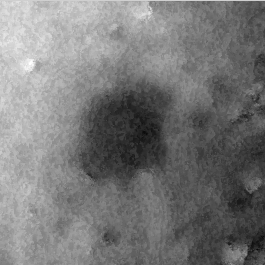}
\label{fig:origDispMap}}
\subfigure[]{\includegraphics[height=1.2in]{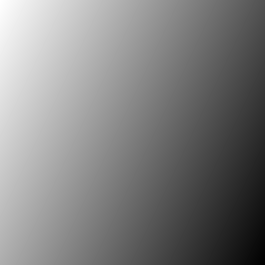}
\label{fig:referencePlane}}
\subfigure[]{\includegraphics[height=1.2in]{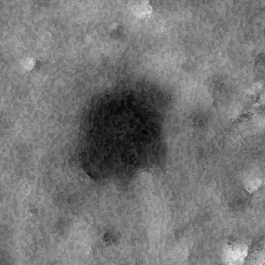}
\label{fig:correctedDispMap}}
\end{center}
\caption{Disparity map and its tilt correction: (a) Cropped stereo fundus photographs centered at ONH; (b) Shape estimate of the optic nerve represented as a grayscale map, where darker pixels indicate points further away from the camera; (c) Reference plane by fitting an orthogonal regression to the optic disc margin; (d) Disparity map after tilt correction by aligning the reference plane horizontally.}
\label{fig:tiltCorrect}
\end{figure}

The original stereo photographs ($3008\times{}2000$ pixels) were manually cropped to images of $651\times{}691$ pixels centered at the optic disc. Disparity maps were obtained by our multi-scale stereo matching algorithm (Fig.~\ref{fig:tiltCorrect}). To correct the tilts of the ONH caused by misalignment of the camera optical axis with respect to the eye’s optical axis, a reference plane is placed by fitting an orthogonal regression to the optic disc margin. The optic disc is assumed to locate at the center of the disparity map with a constant diameter. The retinal surface was flattened by aligning the reference plane horizontally.

After tilt correction, disparity maps were further cropped to $412\times{412}$ pixels to eliminate edge artifacts at boundaries. Since features learned by SAE are invariant to translation, rotation and scaling, one disparity map was cropped at 5 locations with different translations in order to generate enough samples to train SAE. For each location, the cropped map was rotated at 7 different angles, ranging from [-30, 30] degree with 10 degree interval, resulting in the total number of synthesized samples being 204085 $(1444\times{35}+4387\times{35})$. Left eye maps were flipped horizontally to match the shape of right eye maps. Finally all maps were down-sampled to $25\times{}25$ pixels and rescaled within an intensity range of [0,1].


\subsection{ONH Shape Features Learned by SAE}

SAE was trained using both two-layer and three-layer structure similar to the one shown in Fig.~\ref{fig:pretraining} and Fig.~\ref{fig:finetune}. The weight decay parameter $\lambda$ in Eq.~\ref{Eq:cost} is set to 0.003, the weight of the sparsity penalty term $\beta$ is set to 3.0, and the desired level of sparsity $\rho$ is set to 0.1 \cite{AndrewNg:Selftaught}.

\begin{figure} 
\centering
$\begin{array}{c c c}
\includegraphics[height=0.8in]{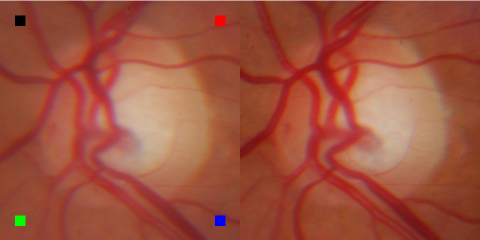}
& \includegraphics[height=0.8in]{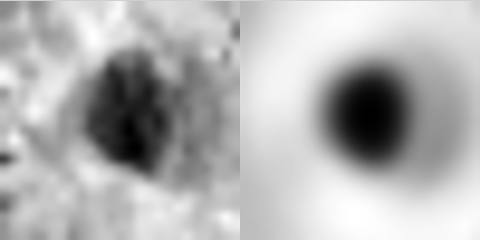}
& \includegraphics[height=0.8in]{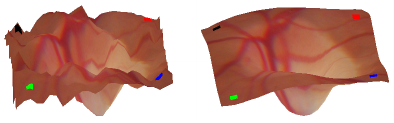}\\
\includegraphics[height=0.8in]{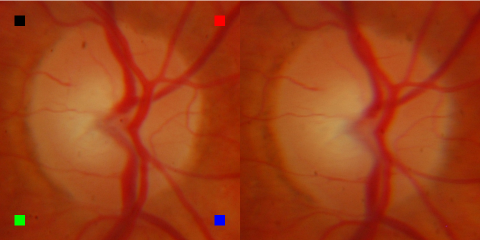}
& \includegraphics[height=0.8in]{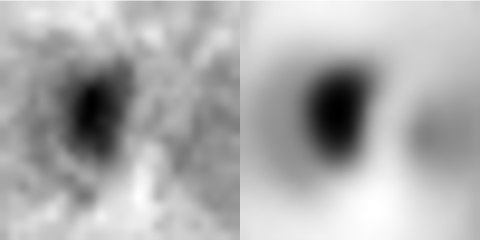}
& \includegraphics[height=0.8in]{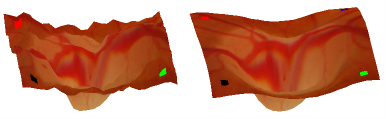}\\
\includegraphics[height=0.8in]{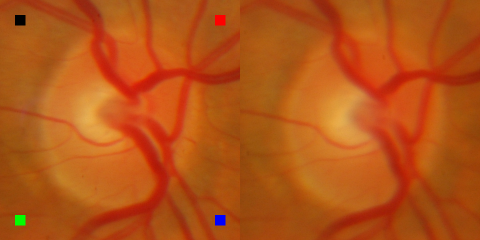}
& \includegraphics[height=0.8in]{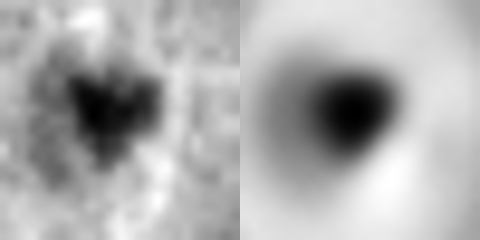}
& \includegraphics[height=0.75in]{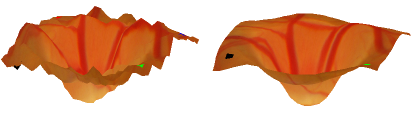}\\
(a) & (b) & (c)
\end{array}$
\caption{Original and reconstructed ONH shape from a two-layer SAE with 300 hidden units for the first layer and 100 units for the second layer: (a) Stereo pair centered at ONH; (b) Original disparity map from stereo matching and reconstructed disparity map from SAE; (c) 3D shape visualization by wrapping the reference (left) fundus image onto the topography described by the original disparity map and the reconstructed one. The orientation of the view angle is indicated by colored dots at four corners.}
\label{fig:dispMap}
\end{figure}

Three disparity maps and those reconstructed maps from SAE are compared in Fig.~\ref{fig:dispMap}. The recovered maps retained major shape variations due to anatomical structure differences of ONH, while high frequency noise were removed. Therefore, some variations of SAE can also be utilized for image denoising. Different combinations of the number of hidden units and the number of layers were tested. The recovered maps are more close to the original ones when the number of hidden units are large, which gives the network more expressing power while increases the risk of over-fitting given limited data for learning.

\begin{figure} 
\centering
		\includegraphics[height=5.3cm]{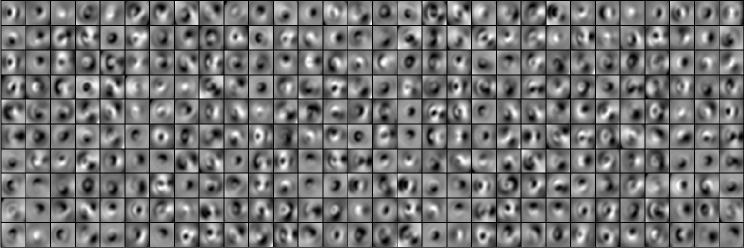}
\caption{300 features learned by the first autoencoder captured localized variability of the cupping of ONH.}
\label{fig:autoencoder1}
\end{figure}

\begin{figure}
\centering
		\includegraphics[height=3.8cm]{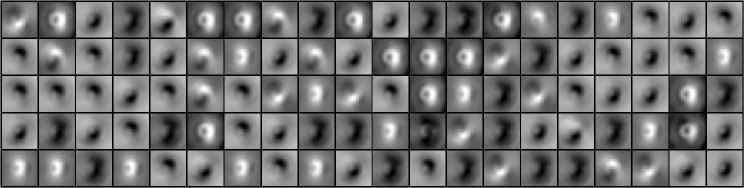}
\caption{100 features learned by the second autoencoder abstract more global and higher level structure of ONH.}
\label{fig:autoencoder2}
\end{figure}

\begin{figure}
\centering
		\includegraphics[height=2.5cm]{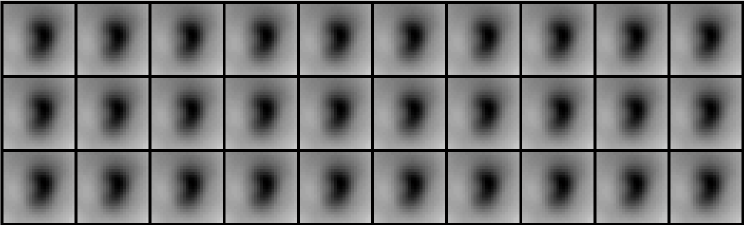}
\caption{The optimal stimulus for the 30 neurons at the third layer confirm that the deep network indeed learned the ONH shape of a right eye.}
\label{fig:autoencoder3}
\end{figure}

The function learned by each layer of the trained SAE can be visualized by displaying the input images that maximally activate each of the hidden units at that layer. For a three-layer network, 300 features learned by the first autoencoder is shown in Fig.~\ref{fig:autoencoder1}, where localized variability of the cupping of ONH is captured; 100 features learned by the second autoencoder is shown in Fig.~\ref{fig:autoencoder2}, where more global and higher level structure is extracted. Some of them resemble outputs from a steerable Gaussian filter \cite{Freeman:PAMI}, which is known for modeling oriented simple cells in the primary visual cortex \cite{Hubel:Wiesel}. Finally, the optimal stimulus for the 30 neurons at the third layer is shown in Fig.~\ref{fig:autoencoder3}, confirming that the deep network indeed learned the ONH shape of a right eye. The learning process demonstrated that the SAE was able to reveal higher level structure of ONH progressively and encode it in a hierarchical feature space.

\begin{figure}
\centering
		\includegraphics[height=4.5cm]{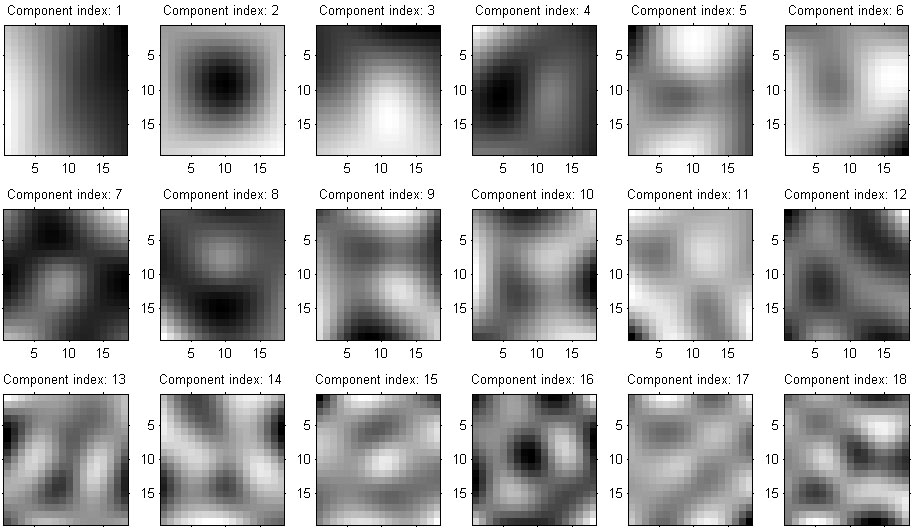}
\caption{First 18 eigen disparity maps from PCA.}
\label{fig:eigenDispMap}
\end{figure}

For comparison, the first 18 eigen disparity maps, i.e. principal components (PCs) of the same dataset, are shown in Fig.~\ref{fig:eigenDispMap}. The top 30 PCs retained 99.96\% of the variability among this dataset. The first few PCs account for the lowest spatial frequencies present in the dataset. The second component shows an average round cup without details learned from a right eye, probably because it is not invariant to translation or rotation. In addition, due to the constraint that each of the component has to be mutually orthogonal, these maps are not spatially localized, lacking an obvious visual interpretation \cite{Seung:Nature}.

\begin{figure} 
\centering
$\begin{array}{c c}
		\includegraphics[height=5.5cm]{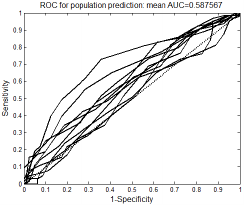} &
		\includegraphics[height=5.5cm]{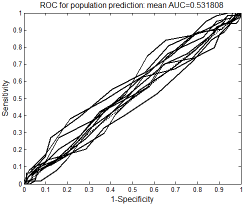}\\
		(a) & (b)
\end{array}$
\caption{The mean area under the receiver operating characteristic curve in predicting the two population with 15 selected features using ten-fold cross-validation: (a) SAE: 0.59 (95\% CI: [0.545, 0.630]); (b) PCA: 0.53 (95\% CI: [0.498, 0.565]).}
\label{fig:ROCcurve}
\end{figure}

Four hundred (300+100) features encoded by the two-layer network for the target dataset, i.e. the Pitcairn dataset, underwent feature selection according to their discriminant power in predicting the two populations. The target dataset was partitioned into 10 subsets. An optimal combination of 15 selected features fed to a K-nearest neighbor classifier (K=30) achieved an average prediction accuracy of $58.4\%$ using ten-fold cross-validation. The mean area under the receiver operating characteristic curve (AUC of ROC) in predicting the two population with selected features is 0.59 (Fig.~\ref{fig:ROCcurve}) with 95\% confidence interval (CI) being [0.545, 0.630]. The mean AUC with 15 selected features from PCA is 0.53 (95\% CI: [0.498, 0.565]).

\begin{figure} 
\centering
$\begin{array}{c c}
		\includegraphics[height=4.0cm]{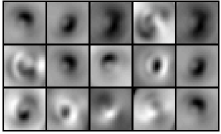} &
		\includegraphics[height=4.0cm]{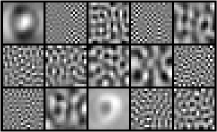}\\
		(a) & (b)
\end{array}$
\caption{Fifteen features selected from: (a) two-layer SAE: feature 4, 6, 8, 9, 11, 12, 14 are from the first layer and feature 1, 2, 3, 5, 7, 10, 13, 15 are from the second layer; (b) PCA components.}
\label{fig:selectedFeatures}
\end{figure}

All of the selected features come from the first and the second autoencoders, as is shown in the left panel of Fig.~\ref{fig:selectedFeatures}. Features extracted at the third autoencoder contain global features of the general ONH shape of a right eye, similar to the fact that the first several eigen disparity maps from PCA corresponds to the mean shape or dominant variation of the input patterns. Higher level features are desirable in distinguishing a disparity map of ONH from other images or patterns, while they are not discriminative enough in identifying subtle localized differences between ONH shape of one population from that of the other. It is entirely dependent upon the specific application of classification tasks, whether a global or a local feature be involved, both of which are readily available from a SAE. Most of the 15 features selected from PCA correspond to PCs with small eigen values, which lack obvious visual interpretations related to the underlying structure.

\subsection{Hierarchical Clustering of Population}

\begin{figure} 
\centering
	$\begin{array}{c}
		\includegraphics[height=1.5cm]{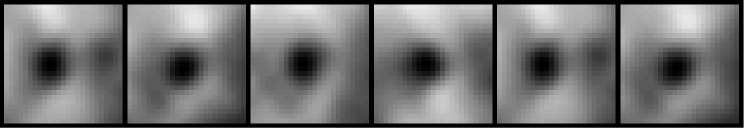}\\ 
		(a)\\
		\includegraphics[height=1.5cm]{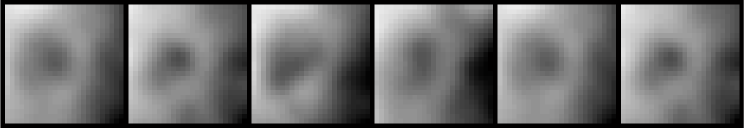}\\ 
		(b)
	\end{array}$
\caption{Two different clusters grouped in terms of ONH shape proximity.}
\label{fig:groups}
\end{figure}

Hierarchical clustering of each individual described by the selected features was conducted for a continuous cluster numbers ranging from 2 to 7. Clustering was repeated 1000 times with random reseeding and the solution with the lowest within-cluster sums of point-to-centroid distances was returned using cosine as distance measure. Fig.~\ref{fig:groups} demonstrated two different clusters grouped in terms of ONH shape proximity. The ONH shape characteristics in top panel are different from that of the bottom panel whereas those within the same cluster is relatively homogeneous. Note that in Fig.~\ref{fig:groups} (a), the six maps are grouped together not only because the overall ONH shapes are very similar, but also because of the similar features present at the superior and inferior sectors, which may indicate important genetic compositions from image based phenotype.

\begin{figure} 
\centering
		\includegraphics[height=2.5cm]{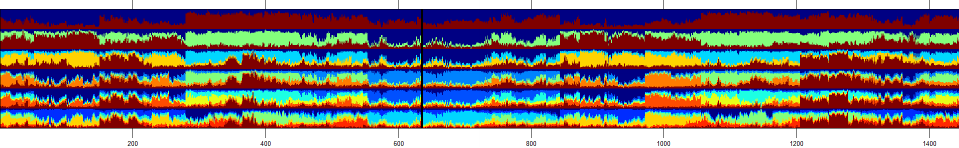}\\
\caption{Estimated population structure at cluster number k=2,3,4,5,6,7. The first 636 optic disc are from inhabitants of Pitcairn and the last 808 from inhabitants of Norfolk (separated by a vertical black line). Cluster membership fractions are indicated by the length of color segments.}
\label{fig:structure}
\end{figure}

The clustering distribution of the ONH shape among the two populations is shown in Fig.~\ref{fig:structure}. Each ONH shape was represented by a column which has $k$ colored segments. The length of each color segment is inversely proportional to the normalized point-to-centroid distances, which indicates fractions of cluster membership. By a comparison of clustering of ONH shape from inhabitants of Pitcairn (the first 636) with those of Norfolk (the last 808), we observed that they share some common population structures, consisting of roughly 5 clusters. Greater within-group variation than expected may be attributed to long range gene flow, and less within-group variation than expected may be attributed to genetic isolation and/or drift \cite{JomonJapan}.

\section{Discussion}

The population structure of 636 Bounty Mutineer descendants and 808 unrelated Norfolk Island inhabitants was analyzed phenotypically using stereo retinal fundus photographs. This dataset offers a unique opportunity to test hypotheses about the utility of image based phenotype analysis using three-dimensional shape variation for automated discovery of population structure. The optic nerve head shape, the complex gene expression we are interested in, consists of complex morphological structures, which pose challenges to explore the phenotypic output expected from admixture on quantitative traits. The three-dimensional shape of the ONH, represented as disparity map derived from fundus photographs using our multi-scale stereo matching algorithm, was quantified into a set of hierarchical vectors for progressively higher level feature representation learned by a deep neural network, which is used to assess the degree of differentiation among clusters of the unique population. Individuals described by a subset of selected features are grouped into $k$ continuous clusters by hierarchical clustering, and cluster membership fractions are determined according to the distance of centroids of each cluster.

This analysis examines relationships between and within groups, and reflects patterns of population structure and degree of morphological variation. Analysis of the phenotypic ONH shape would help further understanding its developmental mechanisms by which the phenotypic variation is expressed, potentially leading to new genetic risk factors for glaucoma and other eye diseases.

Due to a variety of factors involved in the admixture processes, deviations from the expected pattern may be detected, indicating complicated and integrated shape changes which cannot be exclusively explained by a gene flow model. We have conducted a twin study on this phenotype and results show it is largely under genetic control. This may also provide us a reference as how the genetic composition of the traits is likely to differ in isolated population compared with twin samples drawn from the general population.

A limitation of this approach is that both the unlabeled dataset and the labeled dataset may not be large enough for fully exploring the expressive power of the deep neural network. One of the great advantages of such a network is its ability to discover potential relationships or linkable information from large scale dataset, which is also the very property that is highly desirable in field such as Genome-Wide Association Studies (GWAS), especially when big data becomes available. Compared with the parameter space associated with the network, we would expect better results with more data available. The approach we proposed in this study, i.e. automatic three-dimensional ONH shape estimate from low cost fundus photographs, identification of high-dimensional phenotypes from images, quantitative analysis of complex phenotypes from retinal morphology, made more complete study on large dataset possible.

A potential next step would be to examine the entire pedigree of this dataset and analyze the influence of admixture effects on traits and the relationship between each individual in the pedigree for further identification of specific groups of people corresponding to a family. This would also require a larger sample size than what we have for this preliminary study. Comparison of the population structure from Pitcairn and Norfolk island has indicated influence of admixture on the genetic architecture of traits, which may help gene discovery of ONH-related eye diseases.

In summary, population structure of inbred populations, and complex gene expression variation can be discovered from stereo images. Our approach has the potential to allow automated discovery of valid population structure from other stereo images of individuals, including faces.

\section*{Acknowledgments}

This research was supported by the National Eye Institute (R01 EY017066, R01 EY018853), Research to Prevent Blindness, NY, the Department for Veterans Affairs, the American Glaucoma Society, the Carver Center for Macular Degeneration, the Marlene S. and Leonard A. Hadley Glaucoma Research Fund, the American Health Assistance Foundation, Ophthalmic Research Institute of Australia, National Health and Medical Research Council.



\end{document}